\documentclass[conference]{IEEEtran}
\usepackage{graphicx}
\usepackage{amsmath}
\usepackage{amsfonts}
\usepackage{hyperref}
\usepackage{algorithm}
\usepackage{algorithmic}
\usepackage{array}
\usepackage{xcolor}

\usepackage[backend=biber, style=ieee, sorting=none]{biblatex}
\title{Complexity Horizons of Compressed Models in Analog Circuit Analysis}

\author{
    \centering
    \IEEEauthorblockN{Pacome Simon Mbonimpa}
    \IEEEauthorblockA{Research Associate\\
    Carnegie Mellon University-Africa\\
    pmbonimp@andrew.cmu.edu}
}

\begin{document}
\maketitle

\begin{abstract}
The deployment of Large Language Models (LLMs) for specialized engineering domains, such as circuit analysis, often faces a trade-off between reasoning accuracy and computational efficiency. Traditional evaluation methods treat model performance as a flat metric, failing to account for the hierarchical nature of engineering knowledge. We propose a performance-aware model compression strategy that utilizes prerequisite graphs to optimize model selection for circuit analysis tasks. By structuring electronics design concepts as Directed Acyclic Graphs (DAGs), we can identify the specific complexity horizons of an LLM's compressed variants' tiers. Our framework introduces an agentic pipeline for generating prerequisite-based datasets and a strategic evaluation engine that dynamically cascades queries across a spectrum of compressed variants of an LLM. This approach allows to select the smallest compressed model, given its conceptual knowledge boundaries in circuit analysis. Experimental results on analog electronics datasets demonstrate that prerequisite graphs provide a granular map of model compression with respect to the performance given circuit analysis complexity.
\\\textbf{Source Code:} \href{https://github.com/pacomesimon/LLM_prereq_graphs_circuit_analysis}{Code Repository} (Demo: \href{https://huggingface.co/spaces/pacomesimon/LLM_prereq_graphs_circuit_analysis}{App})
\end{abstract}


\section{Introduction}
The advancement of Large Language Models (LLMs) has opened new frontiers in automated engineering and scientific reasoning. In the domain of circuit analysis and design, LLMs are increasingly tasked with complex problem-solving, ranging from basic component selection to high-frequency stability analysis. Recent surveys and frameworks, such as those exploring LLMs for Electronic Design Automation (EDA) \cite{xu2025llm4eda, pan2025survey_llm4eda, zhong2023llm4eda}, demonstrate their growing capability in translating natural language specifications into hardware logic and accelerating the design-to-manufacturing workflow. At the same time, prior work in engineering domains has begun to explore the use of domain-adapted small language models and hybrid large--small model pipelines to mitigate the computational cost of deploying state-of-the-art LLMs for every query, showing that carefully specialized or collaboratively orchestrated smaller models can achieve competitive performance in hardware design and industrial automation tasks \cite{shankar2025smallmodelseda, construction2026smalllm, manufacturing2026hybrid}. However, these approaches lack a principled mechanism to determine which model is sufficient for a given conceptual requirement, as they do not explicitly account for the hierarchical and prerequisite-driven structure of engineering knowledge. Many tasks in electronics design follow such a rigid hierarchy: for instance, a model cannot reliably optimize a transient response without first mastering the foundational principles of negative feedback.

Traditional evaluation benchmarks, which rely on flat sets of largely independent questions, 
such as SuperGLUE \cite{wang2019superglue}, MMLU \cite{mmlu_paper}, BIG-bench \cite{srivastava2022beyond}, and HumanEval \cite{humaneval_paper}, 
fail to capture the prerequisite-driven and hierarchical nature of engineering knowledge. 
These benchmarks typically aggregate performance into scalar metrics, offering limited insight into the conceptual dependencies underlying task success. 
Consequently, they provide little guidance for optimizing model deployment through the strategic selection of the smallest capable model at a given conceptual depth. 
To address this limitation, we propose a framework for analyzing model performance and optimizing deployment using Prerequisite Graphs, which explicitly encode the dependency structure of domain knowledge.

Our approach shifts the paradigm from general-purpose evaluation to domain-specific capability mapping. By modeling electronics design concepts as Directed Acyclic Graphs (DAGs), we create a structured environment where each node represents a task with inherited conceptual complexity. This structure allows us to identify the "Complexity Horizon" of various model tiers, telling us the conceptual limit for each model variant's knowledge and also providing a roadmap for efficient model  compression selection in  engineering workflows.

The key contributions of this work are:
\begin{enumerate}
    \item \textbf{Agentic Prerequisite Graph Generation:} A two-phase pipeline using LLM agents to blueprint electronics conceptual hierarchies and translate them into structured, prerequisite-aware datasets.
    \item \textbf{Performance-Aware Strategic Engine:} A Depth-First Search (DFS) based evaluation engine that navigates prerequisite graphs and manages model cascades, optimizing the model-task fit based on real-time performance boundaries.
    \item \textbf{Conceptual Intelligence Delta Analysis:} A mathematical framework for visualizing model capability boundaries through radial graph layouts and tag-set intersections, enabling granular, performance-aware model compression selection.
\end{enumerate}

The remainder of this paper is organized as follows: Section~\ref{sec:lit_rev} reviews related work in LLM engineering reasoning and model compression; Section~\ref{sec:approach} details the Prerequisite Graph architecture; Section~\ref{sec:evaluation} presents the mathematical foundations for analysis; Section~\ref{sec:discussion} discusses experimental findings; and Section~\ref{sec:conclusion} concludes with future directions.

\section{Literature Review}
\label{sec:lit_rev}
 This section reviews current trends in LLM benchmarking, tool-calling evaluation, and the emerging field of hierarchical task modeling.

\subsection{LLM Benchmarking and Selection For Engineering Tasks}
Initial LLM evaluation relied heavily on flat, static benchmarks such as SuperGLUE \cite{wang2019superglue} (evaluating general language understanding), MMLU \cite{mmlu_paper} (assessing multitask academic knowledge), BIG-bench \cite{srivastava2022beyond} (quantifying broad reasoning capabilities), and HumanEval \cite{humaneval_paper} (measuring functional code generation). While effective for isolated queries, these benchmarks fail to capture the stateful, prerequisite-driven nature of real-world workflows. Similarly, recent efforts in domain-adapted small language models and hybrid pipelines remain largely task-centric. For instance, \cite{shankar2025smallmodelseda} evaluates agentic pipeline success rates and costs, \cite{construction2026smalllm} relies on dataset-level accuracy and scaling, and \cite{manufacturing2026hybrid} measures system workflow completion and latency. By treating tasks as independent units and aggregating performance into scalar metrics, these approaches provide limited insight into underlying conceptual dependencies. Consequently, they fail to reveal capability boundaries or support the principled selection of the smallest sufficient model for complex domains like circuit analysis.

\subsection{Tool-Calling and Agentic Evaluation}
As LLMs are increasingly deployed as agents capable of interacting with external tools, benchmarks such as ToolBench~\cite{toolbench_paper} and AgentBench~\cite{agentbench_paper} have been proposed to evaluate their ability to plan, select, and invoke tools in multi-step scenarios. In parallel, recent works have explored tool-augmented and agentic pipelines for engineering tasks, including circuit analysis and hardware design, where LLMs iteratively generate designs, invoke simulators, and refine outputs based on feedback \cite{shankar2025smallmodelseda, manufacturing2026hybrid}. These frameworks typically evaluate performance through step-wise correctness, task success rates, or end-to-end workflow completion. However, such evaluation protocols largely assume linear or weakly structured task trajectories and fail to capture the complex, non-linear prerequisite dependencies that characterize professional engineering workflows. Consequently, they provide limited insight into the conceptual boundaries of model capabilities or how these boundaries impact tool-augmented reasoning and model selection.

\subsection{Hierarchical and Graph-Based Task Modeling}
The concept of modeling tasks as graphs is well-established in classical AI and operations research, but its application to LLM evaluation is relatively recent. Recent studies have explored structured reasoning paradigms such as Chain-of-Thought (CoT) and Tree-of-Thought (ToT), which improve multi-step problem solving by explicitly decomposing reasoning trajectories into intermediate steps and branching search processes \cite{wei2022chainofthought, yao2023treeofthought}. In parallel, emerging works in electronic design automation have incorporated similar reasoning structures into agentic circuit design pipelines, where LLMs iteratively decompose specifications, invoke simulation tools, and refine circuit implementations using structured reasoning traces \cite{shankar2025smallmodelseda, circuitlm2025eda, heart2025llmcircuits}. However, these approaches primarily operate on internal reasoning or execution traces and do not explicitly formalize task dependencies as external, verifiable structures. Our work builds upon these ideas by representing the task space as a Directed Acyclic Graph (DAG) with conceptual inheritance, thereby moving beyond reasoning heuristics toward an explicit model of prerequisite-driven task structure.

\subsection{Model Cascades and Strategic Deployment}
The economic and computational costs of deploying large language models have led to a growing body of work on model cascades, early-exit strategies, and adaptive routing systems that dynamically select the smallest capable model for a given query. Early formulations such as FrugalGPT~\cite{chen2023frugalgpt} introduced LLM cascades that route queries across models of varying size to optimize cost--accuracy trade-offs. Related approaches in adaptive inference, such as speculative decoding and early-exit architectures, further demonstrate that not all inputs require full model capacity~\cite{leviathan2023speculative, xin2020bertexit}. More recent routing-based systems and mixture-of-experts models extend this idea by learning conditional computation pathways that activate only subsets of parameters or specialist models depending on input complexity~\cite{shazeer2017outrageously, fedus2022switch}. However, these cascade mechanisms are typically evaluated using flat accuracy or efficiency metrics, without considering whether routing decisions respect the underlying hierarchical structure of domain knowledge. Prerequisite graph evaluation directly addresses this limitation by introducing a structured evaluation framework that simulates cascade behavior over interdependent task graphs, enabling robustness analysis of model selection policies under explicit conceptual dependencies.

\subsection{Addressing the Identified Gaps}
The reviewed literature highlights a significant gap: the lack of evaluation frameworks that simultaneously account for hierarchical task dependencies and the strategic dynamics of model cascades. Our framework fills this gap by integrating agentic dataset generation with a DFS-driven evolution engine, providing a more realistic simulation of professional agentic workflows.

\section{Approach}
\label{sec:approach}
The Prerequisite Graph framework is composed of two primary components: an automated dataset generation pipeline and a strategic evaluation engine. This section details the logic and implementation specifications of both.

\subsection{Agentic Dataset Generation}
Traditional datasets are often static and lack structural depth. Our approach addresses this by using a multi-agent pipeline to generate hierarchical task graphs.

\textbf{Phase 1: Conceptual Blueprinting.} An LLM agent generates a tree of "concept nodes" $G = (V, E)$. Each node $v \in V$ represents a discrete task (a blueprint for a multi-choice question to be answered by an LLM) and is defined by a schema containing an identifier $ID_v$, a parent reference $P_v$, a description $D_v$, and a set of complexity tags $T_v$. To ensure logical progression, we enforce a conceptual inheritance rule:
\begin{equation}
    T_{child} = T_{parent} \cup \{T_{unique_1}, T_{unique_2}\}
\end{equation}
where $T_{unique}$ represent 1-2 complexity tags specific to the child node (e.g., if the parent is \texttt{\{"Analog Circuits", "Signal Processing", "Electronic Components", "Transistors", "Amplifiers", "Oscillators"\}}, the child might be \texttt{\{"Analog Circuits", "Signal Processing", "Electronic Components", "Transistors", "Amplifiers", "Oscillators", "Feedback Loops", "Phase Margin Analysis"\}}).

\textbf{Phase 2: Tool-Calling MCQ Translation.} A separate agent iterates over each node $v$. It transforms the description $D_v$ and tags $T_v$ into a Multiple-Choice Question (MCQ) with four options. The ground truth is captured as a structured tool call response, where the agent must provide the \texttt{correct\_idx} $\in \{0, 1, 2, 3\}$.

\subsection{Compression Level Evaluation Engine}
The Strategic Evaluation Engine simulates a model cascade across the generated task graph using a Depth-First Search (DFS) traversal.

\textbf{Initialization.} Let $M = [m_1, m_2, \dots, m_k]$ be a list of compressed LLMs variants sorted by increasing capability (e.g., $m_1 = \text{nano: the smallest / most compressed}$, $m_k = \text{pro: the largest / least compressed}$).

\textbf{DFS State Machine.} The traversal starts from all root nodes ($P_v = \text{null}$). For each node $v$, the engine maintains the following states:
\begin{enumerate}
    \item \textbf{Upgrade Trigger}: The engine tracks $PF_v$, the cumulative path failures from the root to node $v$. If $PF_v > \text{Threshold}$, the \texttt{model\_index} is incremented, effectively upgrading the system to a larger model $m_{i+1}$.
    \item \textbf{State Persistence}: Once a branch is upgraded, all subsequent descendants on that branch are evaluated using the new model $m_{i+1}$ to maintain consistency.
    \item \textbf{Pruning}: If the most capable model $m_k$ fails to solve node $v$, the node is marked as \texttt{failed\_all}. Consequently, all descendants of $v$ are pruned and marked as \texttt{skipped}, as success on the parent is a prerequisite for descendant tasks.
\end{enumerate}

The complete evaluation logic is summarized in Algorithm~\ref{alg:prereq_eval}.

\begin{algorithm}[H]
    \caption{Strategic Prerequisite Graph Evaluation Traversal}
    \label{alg:prereq_eval}
    \begin{algorithmic}[1]
    \STATE \textbf{Input:} DAG $G$, Model List $M = [m_1, \dots, m_k]$, Threshold $T$
    \STATE \textbf{Function} EvaluateNode($v$, $m_{idx}$, $path\_fails$):
    \IF{$v$ is visited} \RETURN \ENDIF
    \STATE $result \gets$ Query($m_{idx}$, $v$)
    \IF{$result$ is Correct}
        \STATE Mark $v$ as Success
        \FOR{each child $c$ of $v$}
            \STATE EvaluateNode($c$, $m_{idx}$, $path\_fails$)
        \ENDFOR
    \ELSE
        \STATE $path\_fails \gets path\_fails + 1$
        \IF{$path\_fails > T$ \AND $m_{idx} < k$}
            \STATE EvaluateNode($v$, $m_{idx} + 1$, $path\_fails$)
        \ELSE
            \STATE Mark $v$ as Fail
            \STATE Mark all descendants of $v$ as Skipped
        \ENDIF
    \ENDIF
    \end{algorithmic}
\end{algorithm}

\section{Evaluation and Visual Analysis}
\label{sec:evaluation}

\subsection{Evaluation Strategy and Dependency Analysis}
\begin{figure}[htbp] 
    \centering
    \includegraphics[width=\linewidth]{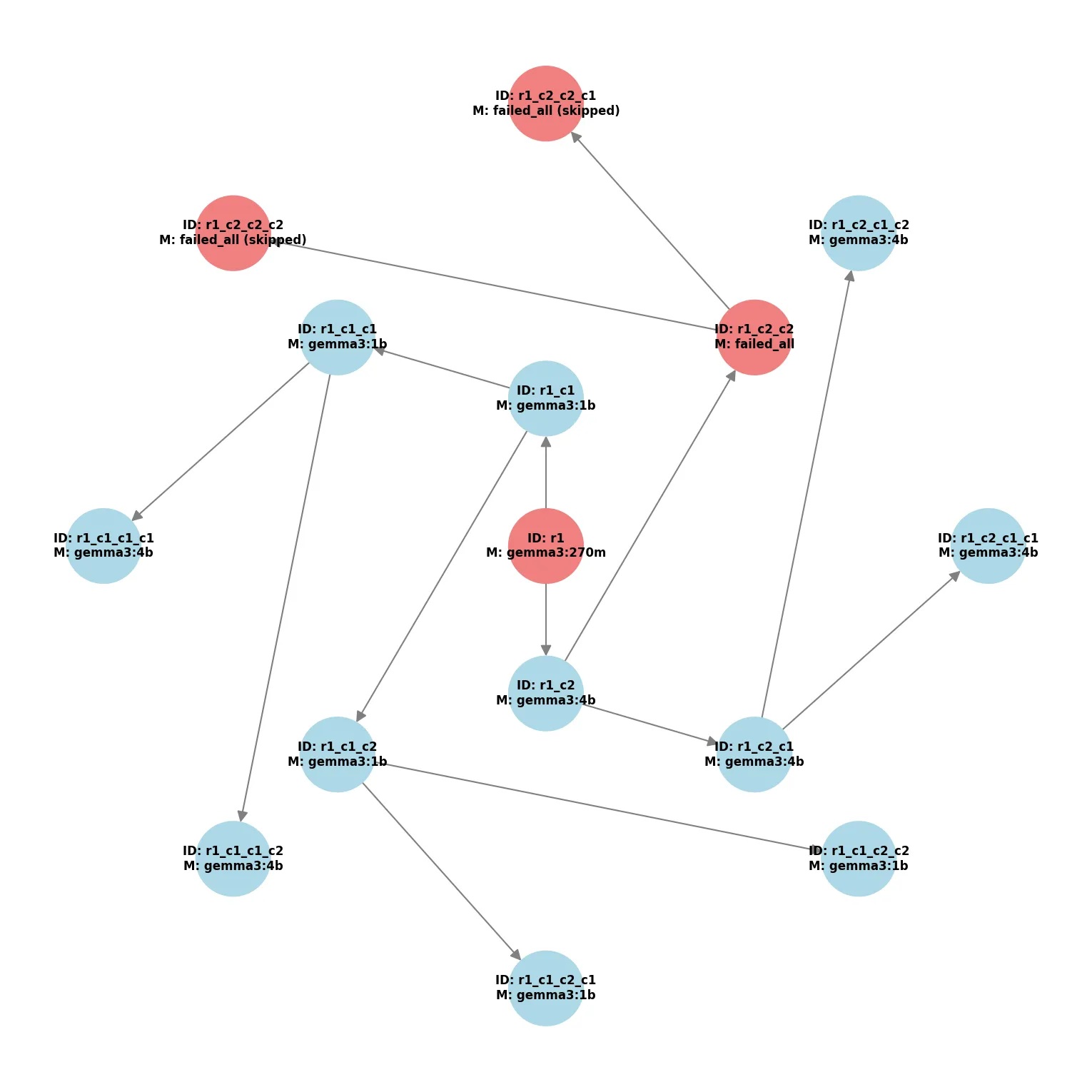}
    \caption{Dependency graph visualization representing the evaluation flow, model transitions, and complexity horizons within the analog electronics dataset. 4 Questions (marked in red) were not answered correctly, 1 being the easiest (failed by the most compressed variant, \texttt{gemma3:270m}). With a path failure threshold of 1, the system automatically transitioned to the next model variant in the list, \texttt{gemma3:1b}, for subsequent questions in the hierarchy. As the graph depth increased, the largest model variant, \texttt{gemma3:4b}, was engaged for more advanced topics. However, once it failed one question, the graph was pruned, skipping other 2 more complex question, with an assumption that this variant won't be able to answer them.}
    \label{fig:dependency_graph}
\end{figure}

We implemented an evaluation strategy using a dependency graph, which serves as a structured dataset of multiple-choice questions about analog electronics. 
As shown in Figure \ref{fig:dependency_graph}, we utilize radial graph layouts to represent the hierarchical status of the task DAG. The nodes are grouped by their Breadth-First Search (BFS) distance from the roots. These groups are mapped to concentric shells using a circular layout algorithm. Nodes are colored based on their outcome (Success: Blue, Fail/Skip: Red), providing an immediate visual representation of the "success depth" of a model cascade.

The root node, representing the most fundamental question, asked: \textit{"Which component in analog circuits is primarily used to amplify weak signals?"} with the options \texttt{["0. Resistor", "1. Transistor", "2. Capacitor", "3. Oscillator"]} and the correct answer being 1.

The evaluation utilized a model cascade consisting of compressed variants from the Gemma family, obtained through large-scale pretraining and distillation-based scaling, including \texttt{gemma3:270m}, \texttt{gemma3:1b}, and \texttt{gemma3:4b}~\cite{gemma2024report, hinton2015distilling}.
In the initial phase, the smallest model variant, \texttt{gemma3:270m}, failed the root question by providing answer 0. Given a path failure threshold of 1, the system automatically transitioned to the next larger model variant in the list, \texttt{gemma3:1b}, for subsequent questions in the hierarchy.

The \texttt{gemma3:1b} variant demonstrated sufficient reasoning capabilities to answer questions tagged with complex concepts such as \textit{Advanced Modulation Techniques}, \textit{Analog-to-Digital Conversion}, \textit{Feedback Systems}, \textit{Mixed-Signal Systems}, \textit{Noise Analysis}, and \textit{Nonlinear Dynamics}. As the graph depth increased, the largest model, \texttt{gemma3:4b}, was engaged for more advanced topics. This model successfully navigated questions involving \textit{Active Filters \& Networks}, \textit{Analog Filter Design}, \textit{Bipolar Junction Transistors}, \textit{Feedback Loops}, \textit{High-Frequency Response Analysis}, \textit{Mixed Signal Design}, \textit{Noise Immunity}, \textit{Non-Linear Circuit Design}, \textit{Nonlinear Circuit Analysis}, \textit{Nonlinear Distortion}, \textit{Operational Stability}, and \textit{Phase Margin Analysis}.

However, the evaluation revealed a complexity horizon when the largest variant in the list failed questions associated with \textit{Negative Feedback Systems} and \textit{Noise Reduction Techniques}. Due to the dependency structure, this failure led to the discarding of subsequent questions tagged with \textit{Non-linear Distortion}, \textit{Nonlinear Distortion Analysis}, \textit{Thermal Stability Analysis}, and \textit{Transient Response Optimization}, as their conceptual prerequisites were not met.

To analyze the results of a prerequisite graph evaluation, we provide a formal mathematical framework for visualizing model performance and conceptual boundaries.

\subsection{Tag-Set Intersection Analysis}

The use of tag-set Venn diagrams provides a powerful way to visualize the "Intelligence Delta" between model tiers. By identifying tags that move from $S_{Fail}(m_{nano})$ to $S_{Pass}(m_{pro})$, we can quantify the specific reasoning capabilities gained when we switch to a lesser compressed model. This granular view is essential for choosing the most cost-effective model for a specific set of task requirements.

To identify specific conceptual boundaries of model intelligence, we perform this set-based analysis on the complexity tags.
Let $T$ be the set of all unique tags in the dataset. We categorize tags into sets based on their associated node outcomes:
\begin{itemize}
    \item $S_{Pass}(m_i)$: Tags associated with nodes solved by model $m_i$.
    \item $S_{Fail}(m_i)$: Tags associated with nodes failed by model $m_i$.
    \item $S_{FailedAll}$: Tags where even the most capable model $m_k$ failed.
\end{itemize}
The intersection $S_{Pass}(m_i) \cap S_{Fail}(m_j)$ (where $i > j$) identifies the "Intelligence Delta"—the specific conceptual capabilities gained when upgrading from $m_j$ to $m_i$.

\subsection{The Monotonic Capability Assumption}
For comparative analysis across different evaluation runs, we apply the following formal assumptions:
\begin{enumerate}
    \item \textbf{Success Inheritance}: We assume that if a smaller model solves a node, any larger model in the cascade would also solve it:
    \begin{equation}
        Pass_{Larger} = Pass_{Larger} \cup Pass_{Smaller}
    \end{equation}
    \item \textbf{Nested Failure}: A global failure is defined by the failure of the largest model:
    \begin{equation}
        FailedAll = Failures_{LargestModel}
    \end{equation}
    \item \textbf{Overall Complexity Horizon}: If the best model fails a node, its entire subtree is assumed to be beyond the system's current reach:
    \begin{equation}
        Skipped = Skipped_{Actual} \cup FailedAll
    \end{equation}
\end{enumerate}

These set analysis allow us to  define the "Complexity Horizon" of a model cascade, which is the point at which increasing the complexity of input queries no longer yields success even with the least compressed model.

\subsection{Validation of the Monotonic Capability Assumption}
The validity of the prerequisite graph evaluation relies on the "Monotonic Capability Assumption," which posits that if a smaller model solves a specific conceptual node, any larger model in the cascade would likewise succeed. To analyze this, we compare the tag-set intersections identified with and without this assumption.

\begin{figure}[htbp]
    \centering
    \includegraphics[width=0.9\linewidth]{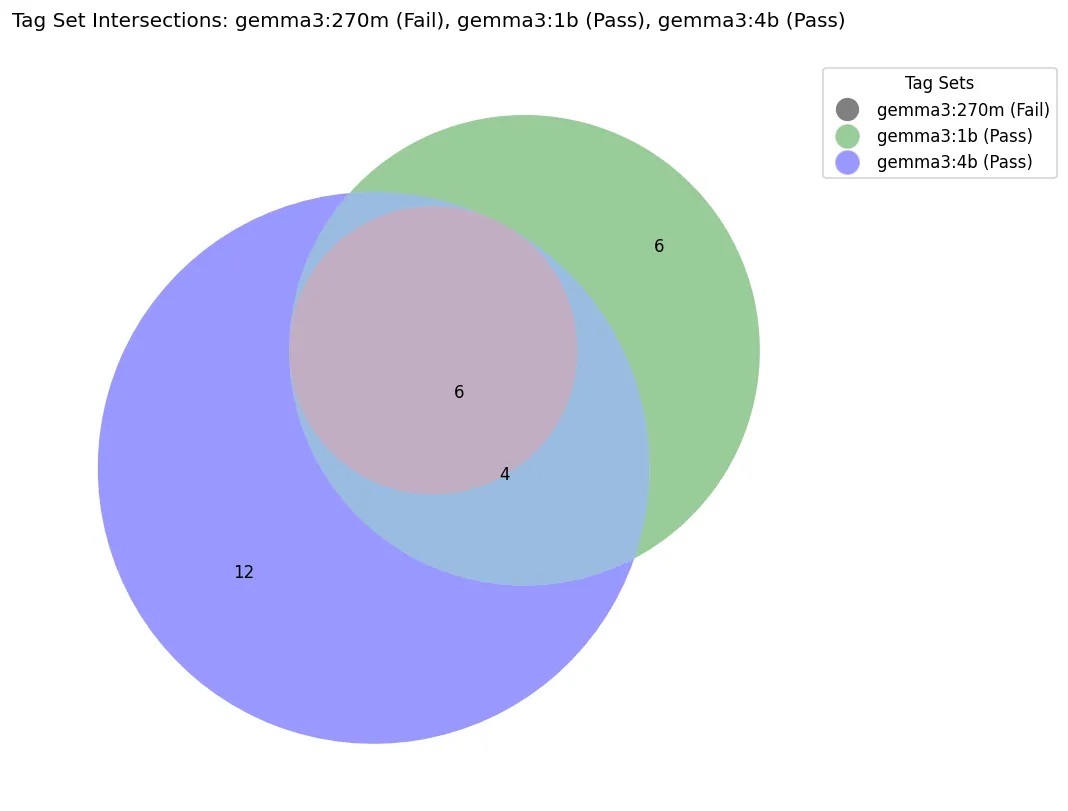}
    \caption{Tag intersections without the monotonic assumption. Pruned graph traversal results in some tags appearing exclusive to smaller models.}
    \label{fig:no_monotonic}
\end{figure}

\begin{figure}[htbp]
    \centering
    \includegraphics[width=0.9\linewidth]{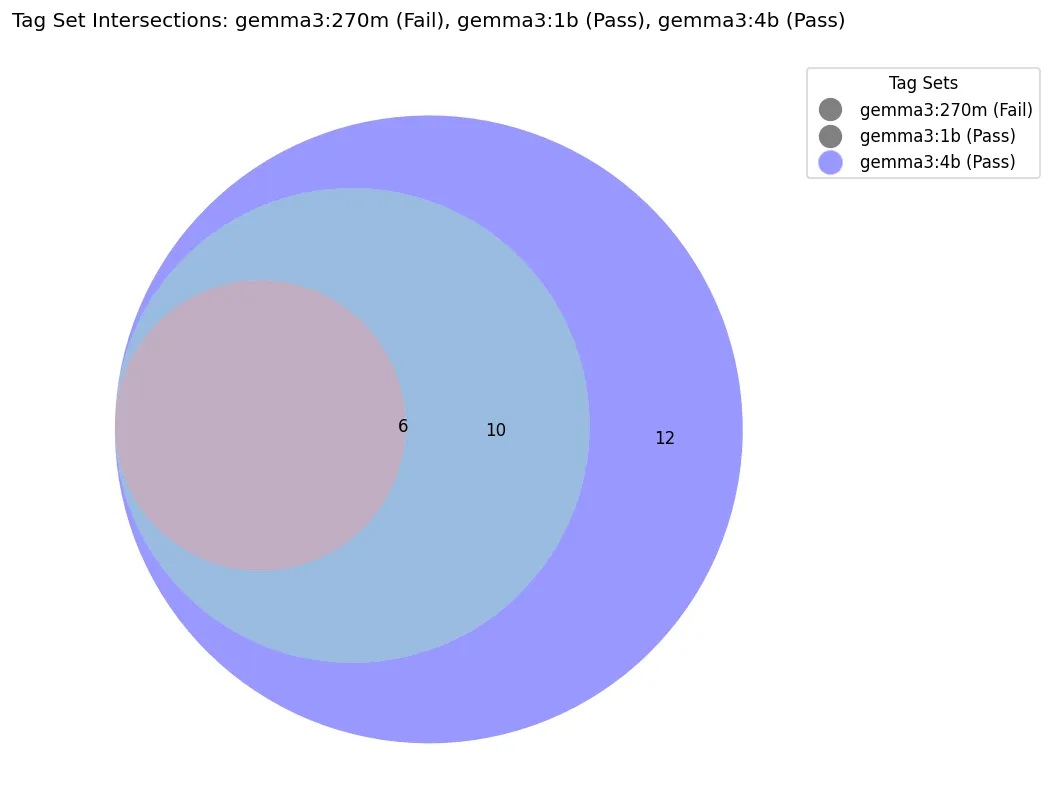}
    \caption{Tag intersections with the monotonic assumption, illustrating the nested capability sets expected in a hierarchical model cascade.}
    \label{fig:monotonic}
\end{figure}

As shown in Figure \ref{fig:no_monotonic}, without the assumption, the \texttt{gemma3:1b} model appears to solve six tags that are not present in the \texttt{gemma3:4b} set: \textit{Advanced Modulation Techniques}, \textit{Analog-to-Digital Conversion}, \textit{Feedback Systems}, \textit{Mixed-Signal Systems}, \textit{Noise Analysis}, and \textit{Nonlinear Dynamics}. This occurs because the DFS traversal reached these leaf nodes while the \texttt{gemma3:1b} model was active, and the system did not need to escalate to the \texttt{gemma3:4b} model.

However, intuitively, these six tags represent concepts that a larger model should handle with ease. This is supported by the fact that the \texttt{gemma3:4b} model successfully answered questions with significantly higher complexity tags, such as \textit{Active Filters \& Networks}, \textit{Analog Filter Design}, \textit{Bipolar Junction Transistors}, \textit{Feedback Loops}, \textit{High-Frequency Response Analysis}, \textit{Mixed Signal Design}, \textit{Noise Immunity}, \textit{Non-Linear Circuit Design}, \textit{Nonlinear Circuit Analysis}, \textit{Nonlinear Distortion}, \textit{Operational Stability}, and \textit{Phase Margin Analysis}. Given that the larger model effectively serves as a more capable version of its smaller counterparts, the monotonic assumption is essential for accurate retrospective capability mapping and identifying the true "Intelligence Delta" of the cascade.

\subsection{Failure Propagation and Conceptual Dependencies}
A critical advantage of the prerequisite graph approach is its ability to identify not only where a model fails but how that failure propagates through a conceptual hierarchy. In our evaluation, the largest model (\texttt{gemma3:4b}) reached its reasoning limit on questions tagged with \textit{Negative Feedback Systems} and \textit{Noise Reduction Techniques}. 

\begin{figure}[htbp]
    \centering
    \includegraphics[width=0.9\linewidth]{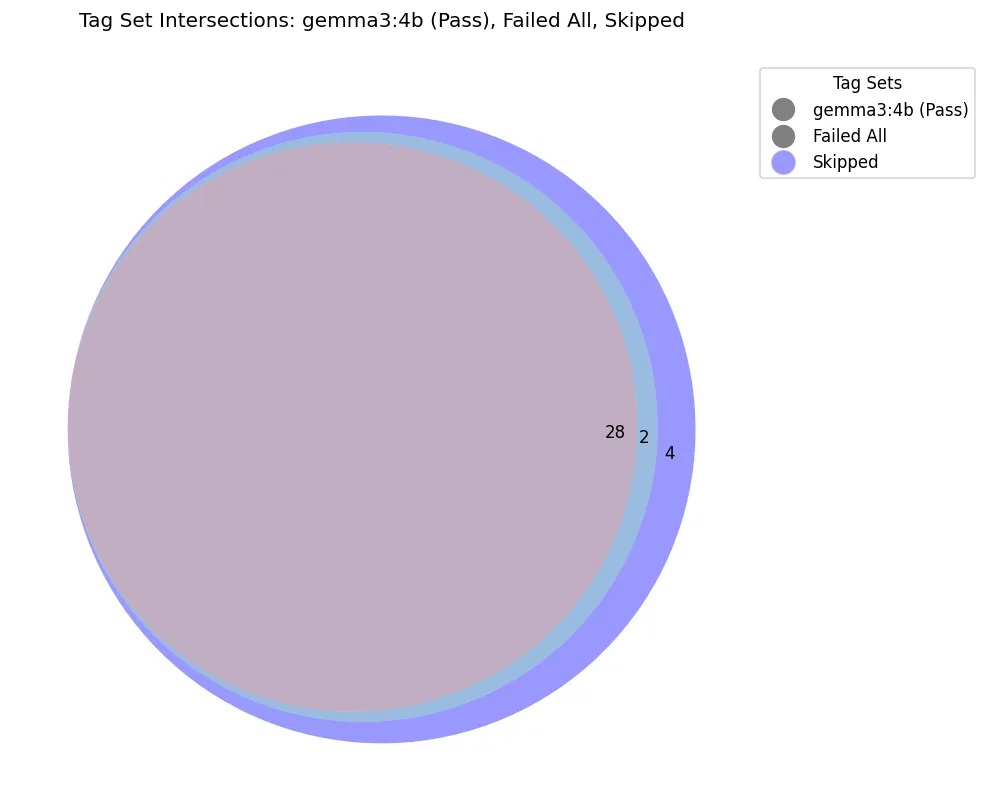}
    \caption{Conceptual boundaries of the most capable model, showing the transition from successful tags to failed prerequisites and their skipped dependencies.}
    \label{fig:failure_propagation}
\end{figure}

The dependency structure of the task graph ensured that once these foundational concepts were failed, the system correctly skipped more advanced topics that relied on them. Specifically, questions associated with \textit{Non-linear Distortion}, \textit{Nonlinear Distortion Analysis}, \textit{Thermal Stability Analysis}, and \textit{Transient Response Optimization} were discarded. This dependency-driven pruning confirms that the evaluation engine effectively mimics the logical progression of electronics design, where mastery of basic feedback and noise reduction is a prerequisite for optimizing transient response or analyzing thermal stability. This approach provides a clear map of the "Conceptual Dead-Ends" where current LLM architectures struggle to maintain logical consistency across interdependent domains.

\section{Discussion}
\label{sec:discussion}
The implementation of the prerequisite graph framework reveals several critical insights into the nature of LLM intelligence and the strategic deployment of model cascades for engineering tasks.

\subsection{The Complexity Horizon}
A key finding of this evaluation is the identification of the "Complexity Horizon"—the point at which a model variant in a cascade fails to solve hierarchical tasks. Unlike flat benchmarks that provide a single accuracy score, our prerequisite graph approach identifies specific conceptual branches where model reasoning breaks down. By analyzing the tags associated with "FailedAll" nodes, researchers can pinpoint precisely which conceptual domains are currently beyond the reach of any model variant in the cascade.

\subsection{Efficacy of the Strategic Upgrade Trigger}
The DFS-based approach, with its path-failure upgrade trigger, demonstrates a more efficient way to utilize model cascades. Traditional cascades often rely on simple intent classification to route queries. In contrast, our performance-aware engine evaluates the model's performance in real-time as it traverses a task graph. The "Success Inheritance" assumption allows for a retrospective analysis of cascade efficiency, suggesting that many tasks currently routed to larger models could be successfully handled by smaller models if the hierarchical context is preserved.

\subsection{Limitations and Constraints}
Despite its advantages, this framework is subject to the limitations of its agentic generation pipeline. The quality of the evaluation is dependent on the "Phase 1" blueprinting agent's ability to create a truly logical and diverse task graph. Furthermore, the "Monotonic Capability Assumption" serves as a useful analytical tool but may not always hold true in practice, as larger models can occasionally exhibit "inverse scaling" \cite{mckenzie2024inversescalingbiggerisnt} on specific, niche tasks.

\section{Future Work}
\label{sec:future_work}
While the prerequisite graph framework provides a robust foundation for hierarchical LLM evaluation in circuit analysis, several avenues for future research remain:

\begin{itemize}
    \item \textbf{Dynamic Task Graph Generation}: Future iterations could incorporate "on-the-fly" task generation, where the blueprinting agent adapts the DAG based on the model's real-time performance, creating a more adversarial and personalized evaluation experience.
    \item \textbf{Cross-Cascade Optimization}: We plan to investigate mathematical models for optimizing the "Upgrade Threshold" across different model combinations. This would benefit from the Depth-First Approach to quickly learn and find the optimal "patience" parameter for each model variant, allowing tolerance in case a model may answer correctly the subsequent questions.
    \item \textbf{Human-AI Collaborative Blueprinting}: Incorporating human experts in the "Phase 1" generation process could ensure that the conceptual hierarchies are not only logically consistent but also aligned with high-stakes professional requirements.
    \item \textbf{Expanding the Monotonic Assumption}: We aim to conduct large-scale empirical studies to validate the "Success Inheritance" assumption across a wider variety of model architectures.
\end{itemize}

By addressing these areas, this prerequisite-aware approach can continue to evolve alongside the rapidly advancing capabilities of Large Language Models, ensuring that engineering evaluation remains both relevant and rigorous.

\section{Conclusion}
\label{sec:conclusion}
In this work, we have introduced a novel paradigm for the performance-aware evaluation and compression of Large Language Models using prerequisite graphs. By shifting from flat benchmarks to hierarchical task graphs (DAGs), our approach provides a more realistic and nuanced assessment of model capabilities in specialized engineering domains like circuit analysis.

Our agentic dataset generation pipeline successfully automates the creation of logically structured tasks with inherited conceptual complexity. Furthermore, the Strategic Evaluation Engine demonstrates how DFS-based traversal and dynamic upgrade triggers can be used to optimize model cascades, identifying the "Complexity Horizon" where model scale may not guarantee success. The mathematical foundations for visual analysis—including radial graph layouts and tag-set Venn diagrams—offer developers a granular view of model intelligence boundaries for performance-aware model selection.

The use of prerequisite graphs represents a significant step towards more robust and professional LLM evaluation, providing the tools necessary to understand and optimize model performance in the next generation of engineering agentic systems.

\printbibliography
\end{document}